\documentclass[letterpaper,12pt]{article}\usepackage{graphicx}
\usepackage{latexsym}
\usepackage[left=1in,top=1in,right=1in,bottom=1in]{geometry}
\usepackage{amsmath}
\usepackage{booktabs}
\usepackage{color}
\usepackage{tikz}
\usepackage{graphicx}
\graphicspath{{figures/}}
\usepackage{caption}
\usepackage{subcaption}
\usepackage{amssymb}
\usepackage{float}
\usepackage{longtable}
\usepackage{amsmath}
\usepackage{algorithm} 
\usepackage{algpseudocode}
\usepackage{algorithmicx}
\usetikzlibrary{shapes}
\usetikzlibrary{arrows}
\definecolor{darkblue}{rgb}{0,0.4,0.9}
\definecolor{gray10}{rgb}{0.1,0.1,0.1}
\definecolor{gray20}{rgb}{0.2,0.2,0.2}
\definecolor{gray30}{rgb}{0.3,0.3,0.3}
\definecolor{gray40}{rgb}{0.4,0.4,0.4}
\definecolor{gray60}{rgb}{0.6,0.6,0.6}
\definecolor{gray80}{rgb}{0.8,0.8,0.8}
\definecolor{gray90}{rgb}{0.9,0.9,.9}
\definecolor{gray95}{rgb}{0.95,0.95,.95}
\definecolor{gray96}{rgb}{0.96,0.96,.96}
\definecolor{lgreen} {RGB}{180,210,100}
\definecolor{dblue}  {RGB}{20,66,129}
\definecolor{ddblue} {RGB}{11,36,69}
\definecolor{lred}   {RGB}{220,0,0}
\definecolor{nred}   {RGB}{224,0,0}
\definecolor{norange}{RGB}{230,120,20}
\definecolor{nyellow}{RGB}{255,221,0}
\definecolor{ngreen} {RGB}{98,158,31}
\definecolor{dgreen} {RGB}{78,138,21}
\definecolor{nblue}  {RGB}{28,130,185}
\definecolor{jblue}  {RGB}{20,50,100}
\definecolor{nnyellow}{RGB}{235,200,0}
\definecolor{purple}{RGB}{150, 0, 120}
\definecolor{sgGreen} {RGB}{20, 180, 50}
\definecolor{revised}{rgb}{0,0,0.9}

\newcommand{\openr}{\hbox{${\rm I\kern-.2em R}$}}
\newcommand{\openn}{\hbox{${\rm I\kern-.2em N}$}}

\usepackage{enumitem}
\newcounter{descriptcount}

\setlist[description]{leftmargin=0.25cm,labelindent=0.25cm}

\graphicspath{{figures/}}
\DeclareMathOperator{\E}{\mathbb{E}}

\DeclareMathOperator*{\argmin}{arg\,min}
\DeclareMathOperator*{\expit}{expit}
\DeclareMathOperator*{\logit}{logit}

\usepackage{mathptmx}

\usepackage{graphics}

\usepackage{natbib} 

\begin{document}

\title{The Relative Performance of Ensemble Methods with Deep Convolutional Neural Networks for Image Classification}

\author{
Cheng Ju and Aur\'elien Bibaut and Mark J. van der Laan
}
\date{}
\maketitle

\begin{abstract}
  
Artificial neural networks have been successfully applied to a variety of machine learning tasks, including image recognition, semantic segmentation, and machine translation. However, few studies fully investigated  ensembles of artificial neural networks. In this work, we investigated multiple widely used ensemble methods, including unweighted averaging, majority voting, the Bayes Optimal Classifier, and the (discrete) Super Learner, for image recognition tasks, with deep neural networks as candidate algorithms. We designed several experiments, with the candidate algorithms being the same network structure with different model checkpoints within a single training process, networks with same structure but trained multiple times stochastically, and networks with different structure. In addition, we further studied the over-confidence phenomenon of the neural networks, as well as its impact on the ensemble methods. Across all of our experiments, the Super Learner achieved best performance among all the ensemble methods in this study.

\end{abstract}



\section{Introduction}


Ensemble learning methods train several baseline models, and use some rules to combine them together to make predictions. The ensemble learning methods have gained popularity because of their superior prediction performance in practice. Consider a prediction task with some fixed data generating mechanism.  The performance of a particular learner depends on how effective its searching strategy is in approximating the optimal predictor defined by the true data generating distribution \citep{van2007super}. In theory, the relative performance of various learners will depend on the model assumptions and the true data-generating distribution. In practice, the performance of the learners will depend on the sample size, dimensionality, and the  bias-variance trade-off of the model.
Thus it is generally impossible to know a priori which learner would perform best given the finite sample data set and prediction problem \citep{van2007super}. One widely used method is to use cross-validation to give an ``objective'' and ``honest'' assessment of each learners, and then select the single algorithm that achieves best validation-performance. This is known as the discrete Super Learner selector \citep{van2003unified,van2007super,polley2010super}, which asymptotically performs as well as the best base learner in the library, even as the number of candidates grows polynomial in sample size.

Instead of selecting one algorithm, another approach to guarantee the predictive performance is to compute the optimal convex combination of the base learners. The idea of ensemble learning, which combines predictors instead of selecting a single predictor, is well studied in the literature: \citep{breiman1996stacked} summarized and referred  several related studies \citep{rao1971combining,efron1973combining,rubin1975variance,berger1976combining,green1991james} about the theoretical properties of ensemble learning. Two widely used ensemble techniques are bagging \citep{breiman1996bagging} and boosting \citep{freund1996experiments,freund1997decision,friedman2001greedy}. Bagging uses bootstrap aggregation to reduce the variance for the strong learners, while  boosting algorithms ``boost'' the capacity of the weak learners. \citep{wolpert1992stacked,breiman1996stacked} proposed a linear combination strategy called stacking to ensemble the models. \citep{van2007super} further extended stacked generalization with a cross-validation based optimization framework called Super Learner, which finds the optimal combination of a collection of prediction algorithms by minimizing the cross-validated risk. Recently, the super learner have showed great success in variety of areas, including precision medicine \citep{luedtke2016super}, mortality prediction\citep{pirracchio2015mortality,chambaz2016data}, online learning \citep{benkeser2016online}, and spatial prediction\citep{davies2016optimal}.

In recent years, deep artificial neural networks (ANNs) have led to a series of breakthroughs in a variety of tasks. ANNs have shown great success in almost all  machine learning related challenges across different areas, like computer vision \citep{krizhevsky2012imagenet,szegedy2015going,he2015deep}, machine translation \citep{luong2015effective,cho2014learning}, and social network analysis \citep{perozzi2014deepwalk,grover2016node2vec}. Due to their high capacity/flexibility, deep neural networks usually have high variance and low bias. In practice, model averaging with multiple stochastically trained networks is commonly used to improve the predictive performance. \citep{krizhevsky2012imagenet} won the first place in the image classification challenge of ILSVRC 2012, by averaging 7 CNNs with same structure. \citep{simonyan2014very} won the first place in classification and localization challenge in  ILSVRC 2014 with averaging of multiple deep CNNs. \citep{he2015deep} won the first place using six models of Residual Network with different depth to form an ensemble in ILSVRC 2015. In addition, \citep{he2015deep} also won the ImageNet detection task in ILSVRC 2015 with the ensemble of 3 residual network models. 

However, the behavior of ensemble learning with deep networks is still not well studied and understood. First, most of the neural networks literature focuses mainly on the design of the network structure, and only applies naive averaging ensemble to enhance the performance. To the best of our knowledge, no detailed work investigates, compares and discusses ensemble methods for deep neural networks.  Naive unweighted averaging, which is largely used, is not data-adaptive and thus vulnerable to a ``bad'' library of base learners: it works well for networks with similar structure and comparable performance, but it is sensitive to the presence of excessively biased base learners. This issue could be easily addressed by a cross-validation based data-adaptive ensemble like Bayes Optimal Classifier and  Super Learner. In later sections, we investigate and compare the performance of four commonly used ensemble methods on an image classification task, with deep convolutional neural networks (CNNs) as base learners.

This study mainly focuses on the comparison of ensemble methods of CNNs for image recognition. For readers who are not familiar with deep learning, each CNN could be just treated as a black-box estimator, with an image as input, and outputs the probability vector for each possible class. We refer the interested reader to \citep{lecun2015deep,goodfellow2016deep} for more details about deep learning.

\section{Background}


In this paper, ``algorithm candidate'', ``hypothesis'', and ``base learner'' refer to an individual learner (here a deep CNN) used in an ensemble. The term 'library' refers to the set of the base learners for the ensemble methods.

\subsection{Unweighted Average}

Unweighted averaging is the most common  ensemble approach for neural networks. It takes unweighted average of the output score/probability for all the base learners, and reports it as the predicted score/probability.

Due to the high capacity of deep neural networks, simple unweighted averaging improves the performance substantively. Taking the average of multiple networks reduces the variance, as deep ANNs have high variance and low bias.  If the models are uncorrelated enough, the variance of models could be dramatically reduced by averaging. This idea inspires Random Forest \citep{breiman2001random}, which builds less correlated trees by bootstrapping observations and sampling features.

We could average either directly the score output, or the predicted probability after softmax transformation:
\begin{equation*}
p_{ij} = \text{softmax}(\vec{s_i})[j] = \frac{\vec{s_i}[j]}{\sum_{k=1}^{K}\exp(s_{i}[k])},
  \end{equation*}

where score vector $\vec{s_i}$ is the output from the last layer of the neural network for $i$-th unit, $\vec{s_i}[k]$ is the score corresponding to k-th class/label, and $p_{ij}$ is the predicted probability for unit $i$ in class $j$. It is more reasonable to average after the softmax transformation, as the scores might have varying scales of magnitude across the base learners, as the score output from different network might be in different magnitude. Indeed, adding a constant to scores for all the classes  leaves predicted probability unchanged. In this study, we compared both naive averaging of the scores and averaging of their softmax transformed counterparts (i.e. the probabilities)

Unweighted averaging might be a reasonable ensemble for similar base learners of comparable performance, as the deep learning literature suggests \citep{simonyan2014very,szegedy2015going,he2015deep}. However, when the library contains heterogeneous networks, the naive  unweighted averaging may not be a smart choice. It is  vulnerable to the weaker learners in the library, and sensitive to the over-confident candidate (We will explain further the over-confidence phenomenon in later sections.). A good meta-learner should be intelligent  enough to combine the strength of base learners data-adaptively. Heuristically, some networks might have weak overall prediction strength, but can be good at discriminating certain subclasses (e.g. fine-grained classifier). We hope the meta-learner could combine the strengths of all the base learners, thus yielding a better strategy.

\subsection{Majority Voting}

Majority voting is similar to  unweighted averaging. But instead of averaging over the output probability, it counts the votes of all the predicted labels from the base learners, and makes a final prediction using label with most votes. Or equivalently, it takes an unweighted average using the label from base learners and chooses the label with the largest value.

Compared to naive averaging, majority voting is less sensitive to the output from a single network. However, it would still be dominated if the library contains multiple similar and dependent base learners. Another weakness of majority voting is the loss of information, as it only uses the predicted label.

\citep{kuncheva2003limits} showed  pairwise dependence plays an an important role in majority voting. For image classification,  shallow networks usually give more diverse prediction compared to deeper networks\citep{choromanska2015loss}. Thus we hypothesize majority voting would yield a greater improvement over base learners with a library of shallow networks than with a library of deep networks.

\subsection{Bayes Optimal Classifier}

In a classification problem, it can be shown that the function $f$ of the predictors $\textbf{x}$ that minimizes the misclassification rate $\E I(f(\textbf{x}) \neq y)$ is the so-called Bayes classifier. It is given by
$f(x) = \text{argmax}_{y}P[y |\textbf{x}]$. It fully characterized by the data-generating distribution $P$.

In the Bayesian voting approach, each base learner $h_j$ is viewed as an hypothesis made on the functional form of the conditional distribution of $y$ given $\textbf{x}$. More formally, denoting $S_{train}$ our training sample, and $(\textbf{x}, y)$ a new data-point, we denote $h_j(y|\textbf{x}) = P[y | \textbf{x}, h_j, S_{train}]$. It means the value of  the hypothesis ${h}_j$, which is trained on $S_{train}$,  evaluated at $(y,x)$. The Bayesian voting approach requires a prior distribution that, for each $j$, models the probability $P(h_j)$ that the hypothesis $h_j$ is correct. Using the Bayes rule, one readily obtains that

\begin{equation}
P(y|\textbf{x}, S_{train}) \propto \sum_{h_j} P[y |h_j, \textbf{x}, S_{train}] P[S_{train} | h_j] P[h_j].
\end{equation}

This motivates the definition of the Bayesian Optimal classifier as

\begin{equation}
\text{argmax}_{y}  \sum_{h_j} h_j(y|\textbf{x}) P[S_{train} | h_j] P[h_j].
\end{equation}
Note that $P[S_{train} | h_j] = \prod_{(y,x) \in S_{train}}h_j(y|x)$ is the likelihood of the data under the hypothesis $h_j$. However this quantity might not reflect well the quality of the hypothesis since the likelihood of the training sample is subject to overfitting. To give an ``honest'' estimation, we could split the training data into two  sets, one for model training, and the other  for computing $P[S_{\text{train}}|h]$. For neural networks, a validation set  (distinct from the testing set) is usually set aside only to tune a few hyper-parameters, thus the information in it is not fully exploited. We expect that using such a validation set would provide a good estimation of the likelihood $P[S_{train} | h]$.  Finally, we would assess the model using the untouched testing set.

      The second difficulty in BOC is choosing the prior probability for each hypothesis $p(h_i)$. For simplicity, the prior is usually set to be the uniform distribution \citep{mitchell1997machine}.

\citep{dietterich2000ensemble} observed that, when the sample size is large, one hypothesis typically tends to have a much larger posterior probability than others. We will see in the later section that when the validation set is large, the posterior weight is usually dominated by only one hypothesis (base learner). As the weights are proportional to the likelihood on the validation set, if the weight vector is dominated dominated by a single algorithm, BOC would be the same selector as the discrete Super Learner selector with negative likelihood  loss function \citep{van2007super}.

\subsection{Stacked Generalization}

The idea of stacking was originally proposed in  \citep{wolpert1992stacked}, which concludes stacking works by deducing the biases of the generalizer(s) with respect to a provided learning set. \citep{breiman1996stacked} also studied  stacked regression by using cross-validation to construct the 'good' combination.

Consider a linear stacking for the prediction task. The basic idea of stacking is to 'stack' the predictions $f_1, \cdots, f_m$   by linear combination with weights $a_i$, $i \in 1, \cdots, m$:

$$f_{stacking}(\bold{x}) = \sum_{i=1}^{m}a_i f_i(\bold{x})$$

where the weight vector $a$ is learned by a meta-learner. 

\section{Super Learner: a Cross-validation based Stacking}

Super Learner \citep{van2007super} is an extension of stacking. It is a cross-validation based ensemble framework, which minimizes cross-validated risk for the combination. The original paper \citep{van2007super} demonstrated the finite sample and asymptotic properties of the Super Learner. The literature shows its  application to a wide range of topics, e.g. survival analysis \citep{hothorn2006survival}, clinical trial \citep{sinisi2007super}, and mortality prediction \citep{pirracchio2015mortality}. It combines the base learners by cross-validation. Here is an example of SL with $V$-fold cross-validation with $m$ base learners for binary prediction. We first define the cross-validated loss for $j$-th base learner:

  \begin{equation*}
\text{R}_{CV}^{(j)} =  \sum_{v = 1}^{V}\sum_{i\in val(v)}l \left(y_i, p_{ji}^{-v}\right)
  \end{equation*}

  where $val(v)$ is the set of indices of the observations in the $v$-th fold, and $p_{ji}^{-v}$ is defined as the prediction for the $i$-th observation, from the $j$-th base learner that trained on the whole data except the $v$-th fold. Then we have

  $$
  \text{R}_{CV}(\vec{a}) = \sum_{v = 1}^{V}\sum_{i\in val(v)}l \left( y_i, \sum_{j=1}^{m}a_{j}p_{ji}^{-v} \right)$$

where $\vec{a} = [a_1,\cdots, a_m]$ is the weight vector. The optimal weight vector given by the Super Learner is then

$$\vec{a} = \argmin_{\vec{a}} \text{R}_{CV}(\vec{a}) $$

For simplicity, we consider the binary classification task, which could be easily generalized to multi-class classification and regression. We first study a simple version of the Super Learner with $m$ single  algorithms, using negative (Bernoulli) log-likelihood as loss function:
$$l(y, p) = -[y\log(p) + (1-y)\log(1-p)].$$
Thus the cross-validated loss is:

 $$
  \text{R}_{CV}(\vec{a}) = -\sum_{v = 1}^{V}\sum_{i\in val(v)}[y_i \log(\sum_{j=1}^{m}a_{j}p_{ji}^{-v} ) + (1-y_i) \log(1 - \sum_{j=1}^{m}a_{j}p_{ji}^{-v})]$$
where $p_{ji}^{-v}$ is the predicted probability for  $i$-th unit from $j$-th base learner which is trained on the whole data except $v$-th fold.

In addition, stacking on the logit scale usually gives much better performance in practice. In other words, we use the optimal linear combination before softmax transformation:

  $$
  \text{R}_{CV}(\vec{a}) = \sum_{v = 1}^{V}\sum_{i\in val(v)}l(y_i, \expit(\sum_{j=1}^{m}a_{j}\logit(p_{ji}^{-v})))$$

For $K$-class classification with softmax output like neural networks, we could also ensemble in the score level:

$$p_i^z(\vec{a}) =  - \log(\frac{\exp(\sum_{j=1}^{m}  a_{j} \cdot s_i[j, z])}{\sum_{k=1}^{K}\exp(\sum_{j=1}^{m}  a_{j} \cdot s_i[j, k])})$$

where $p_i^{z}(\vec{a})$ is the ensemble prediction for $i$-th unit and $z$-th class with weight vector $\vec{a}$. $\bold{s}_i$ is an $m$ by $K$ matrix, and $s_i[j, k]$ stands for the score of $j$-th model and $k$-th class.

We can impose restrictions on $a$, such as constraining it to lie in a probability simplex:

$$||a||_1 = 1, a_i \geq 0, \text{for } i = 1,\cdots, m.$$

This would drive the weights of some base learners to zero, which would reduce the variance of the ensemble and make it more interpretable. This constrain is not a necessary condition to achieve the oracle property for SL. In theory, the oracle inequality requires bounded loss function, so the LASSO constraint is highly advisable (e.g. $\sum_j |a_j| < M$, for some fixed $M$). 
In practice, we found imposing large $M$ leads to better practical performance. 

\begin{figure}[ht]
  \centering
    \includegraphics[width=0.8\textwidth]{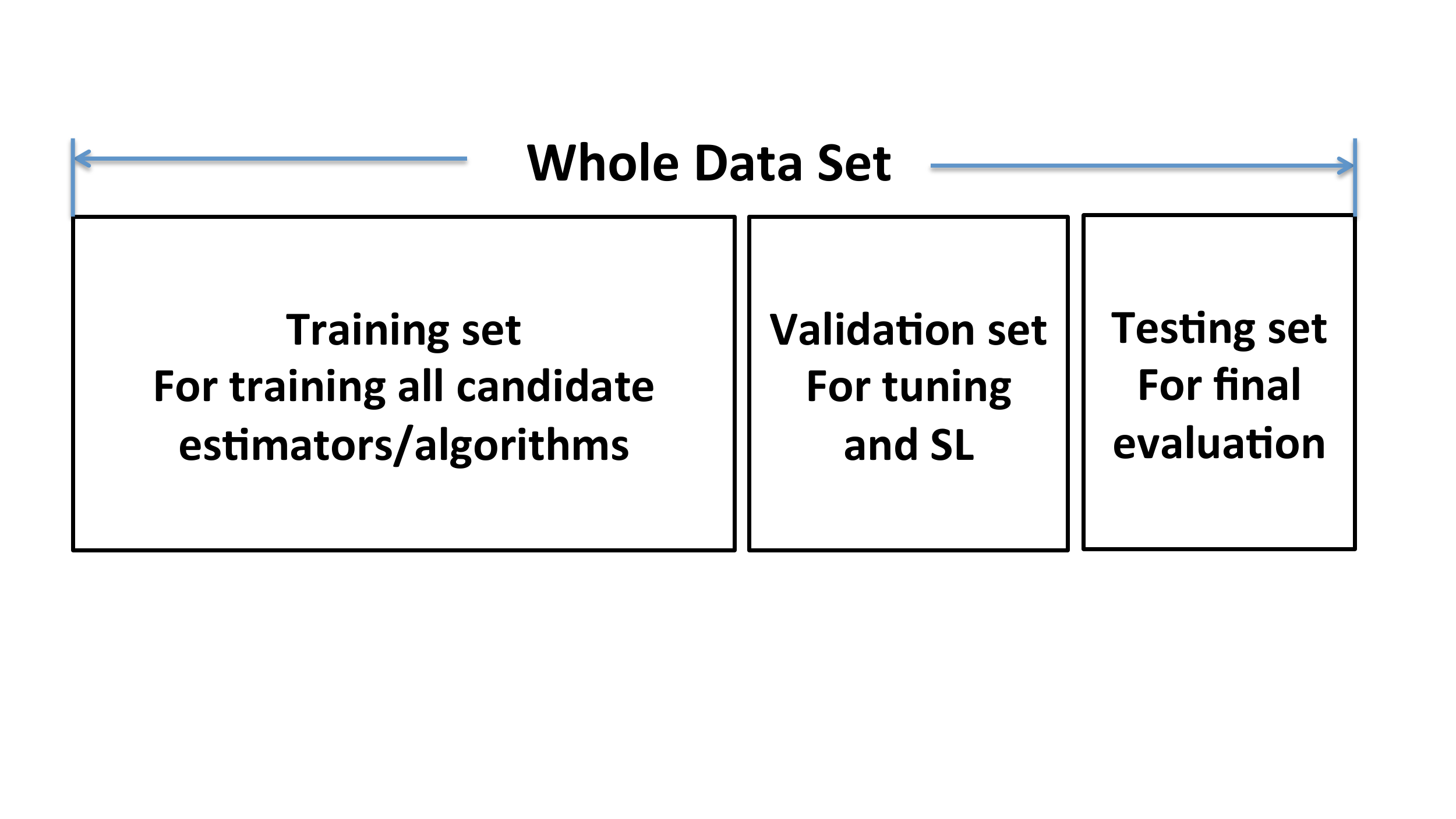}
    \caption{Single Split (Sample Split) Super Learner, which computes the  weights on the validation set.}
    \label{fig:data}
\end{figure}

For small data sets, it is recommended to use cross-validation to compute the optimal ensemble weight vector. However this takes a long time when the data set and the library are large. Usually people just set aside a validation set, instead of cross-validation, to assess and tune the models for deep learning. Similarly, instead of optimizing  the V-fold cross-validated loss, we could optimize on the single-split cross-validation loss instead to get the ensemble weights,  which is so called ``single split (or sample split) Super Learner''. Figure \ref{fig:data} shows the details of this variation of Super Learner. \citep{ju2016propensity} shows the success of such single split Super Learner in three large healthcare databases. In this study, we compute the weights of Super Learner by minimizing the single-split cross-validated loss. This procedure necessitates almost no additional computation: only one forward pass for all validation images and then solving a low-dimensional convex optimization.

\subsection{Super Learner From a Neural Network Perspective}

Lots of neural network structures could be considered as ensemble learning. One of the commonly used regularization methods for deep neural network, dropout \citep{srivastava2014dropout}, randomly removes certain proportion of the activations (the output from the last layer) during the training and uses all the activations in the testing. It could be seen as training multiple base learners and ensemling them during prediction. \citep{veit2016residual} discusses ResNet, a state-of-the-art network structure, could be understood as an exponential ensembles of shallow networks. However, such ensembles might be highly biased, as the meta-learner computes the weights based on the prediction of the base learner (e.g. shallow network) on the training set. These weights might be biased as the base-learners might not make objective prediction on the training set.

In contrast, the Super Learner computes an honest ensemble weight based on the validation set. A validation set is commonly used to train/tune a neural network. However, it is usually only used to select a few tuning parameters (e.g. learning rate,  weight decay). For most image classification data sets, the validation set is  very large in order to make the validation stable.  We thus conjecture that the potential of the validation information has not been fully exploited.

The Super Learner could be considered as a neural network with 1 by 1 convolution over the validation set, with the scores of the base learners as input. It  learns the $1 \times 1 \times m$ kernel either by back-propagation, or through directly solving the convex optimization problem.

\begin{figure}[ht]
  \centering
  \includegraphics[width=0.8\textwidth]{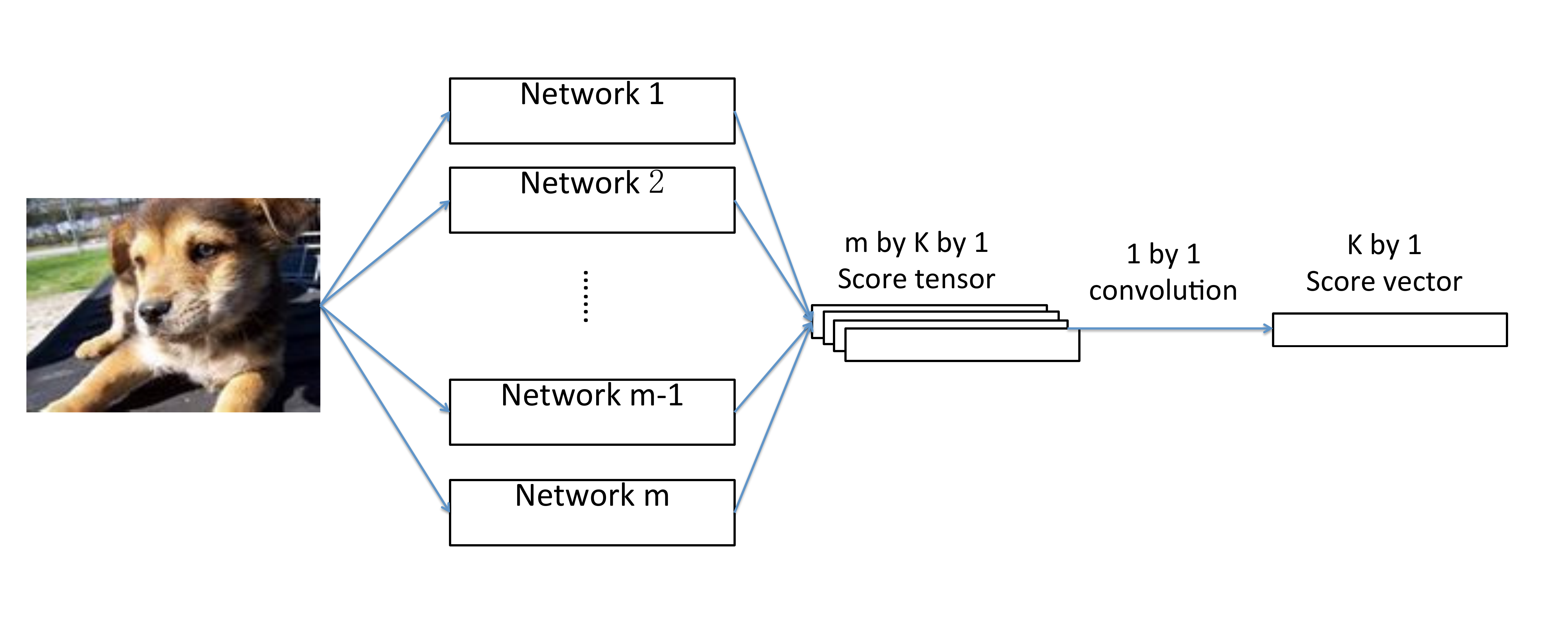}
  \caption{Super Learner from convolution neural network perspective. The base learners are trained in the training set, and 1 by 1 convolutional layer is trained in the validation set. The simple structure of SL avoids the overfitting on the validation set.}
  \label{fig:SL-NoN}
\end{figure}

\section{Experiment}

\subsection{Data}

The CIFAR-10 data set \citep{krizhevsky2009learning} is a widely used benchmark data set for image recognition. It contains  $10$ classes of natural images, with $50,000$ training images and $10,000$ testing images. Each image is an RGB image of size $32 \times 32$. There are $10$ classes in the data set: airplane, automobile, bird, cat, deer, dog, frog, horse, ship, and truck. Each class has  $5000$ images in the training data and $1000$ images in the testing data.

\subsection{Network description}

\subsubsection{Network in Network}

The network in network (NIN) structure \citep{lin2013network} consists of mlpconv (MLP) layers, which use multilayer perceptrons to convolve the input. Each MLP layer is made by one convolution layer with larger kernel size followed by two $1 \times 1$ convolution layer and max pooling layer. In addition, it uses a global average pooling layer as a replacement for the fully connected layers in conventional neural networks.

\begin{figure}[ht]
  \centering
    \includegraphics[width=0.8\textwidth]{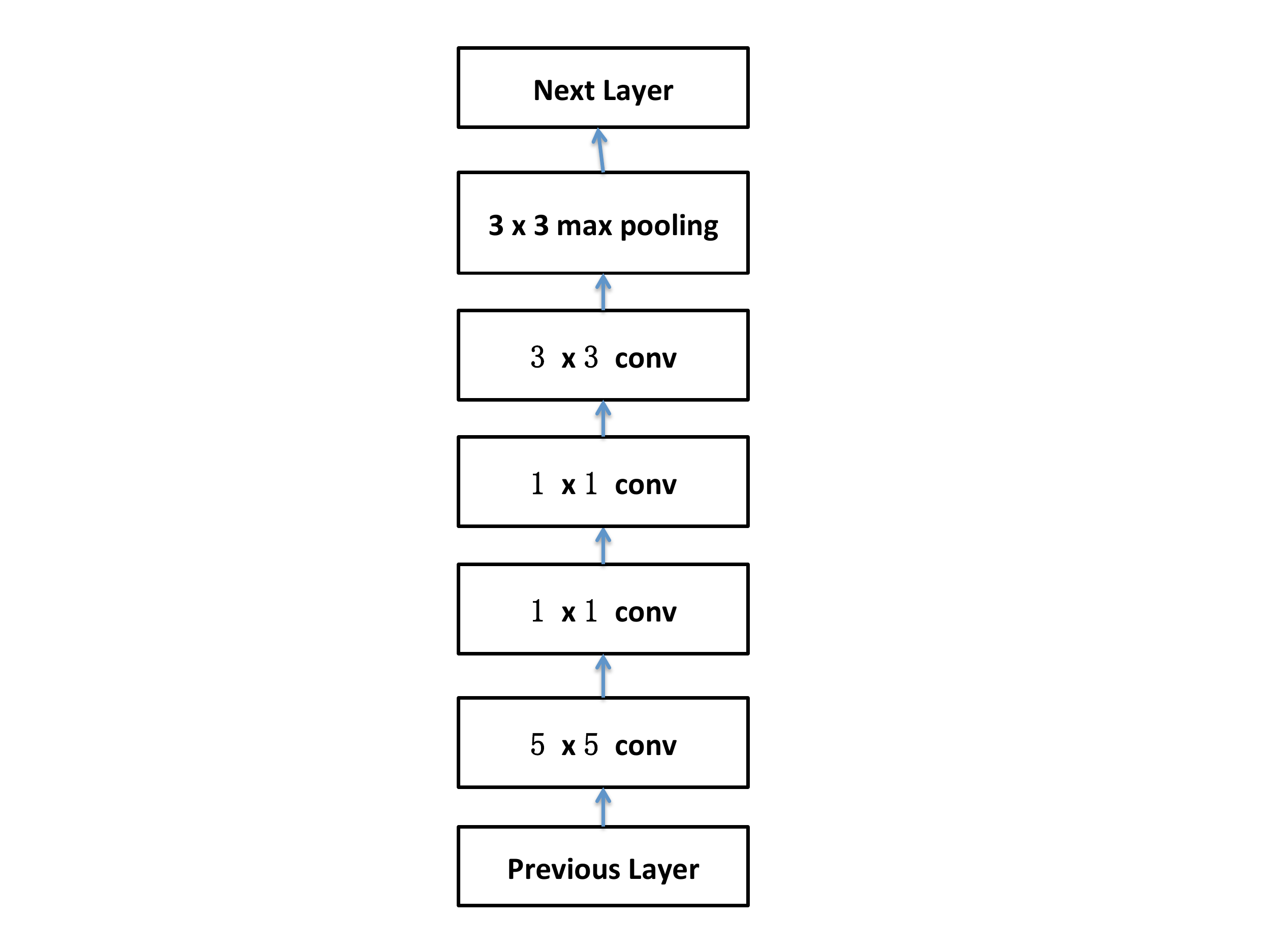}
    \caption{An example of MLP layer in the NIN structure. Notice each convolution are followed by ReLU layer.}
    \label{fig:mlp}
\end{figure}

\subsubsection{GoogLeNet}

GoogLeNet \citep{szegedy2015going} is a deep convolutional neural network architecture based on the inception module, which improved the computational efficiency. In each inception module, a $1 \times 1$ convolution is applied as dimension reduction before expensive large convolutions. Within each inception module, the propagation splits into 4 flows, each with different convolution size, and is then concatenated.

\begin{figure}[ht]
  \centering
    \includegraphics[width=0.8\textwidth]{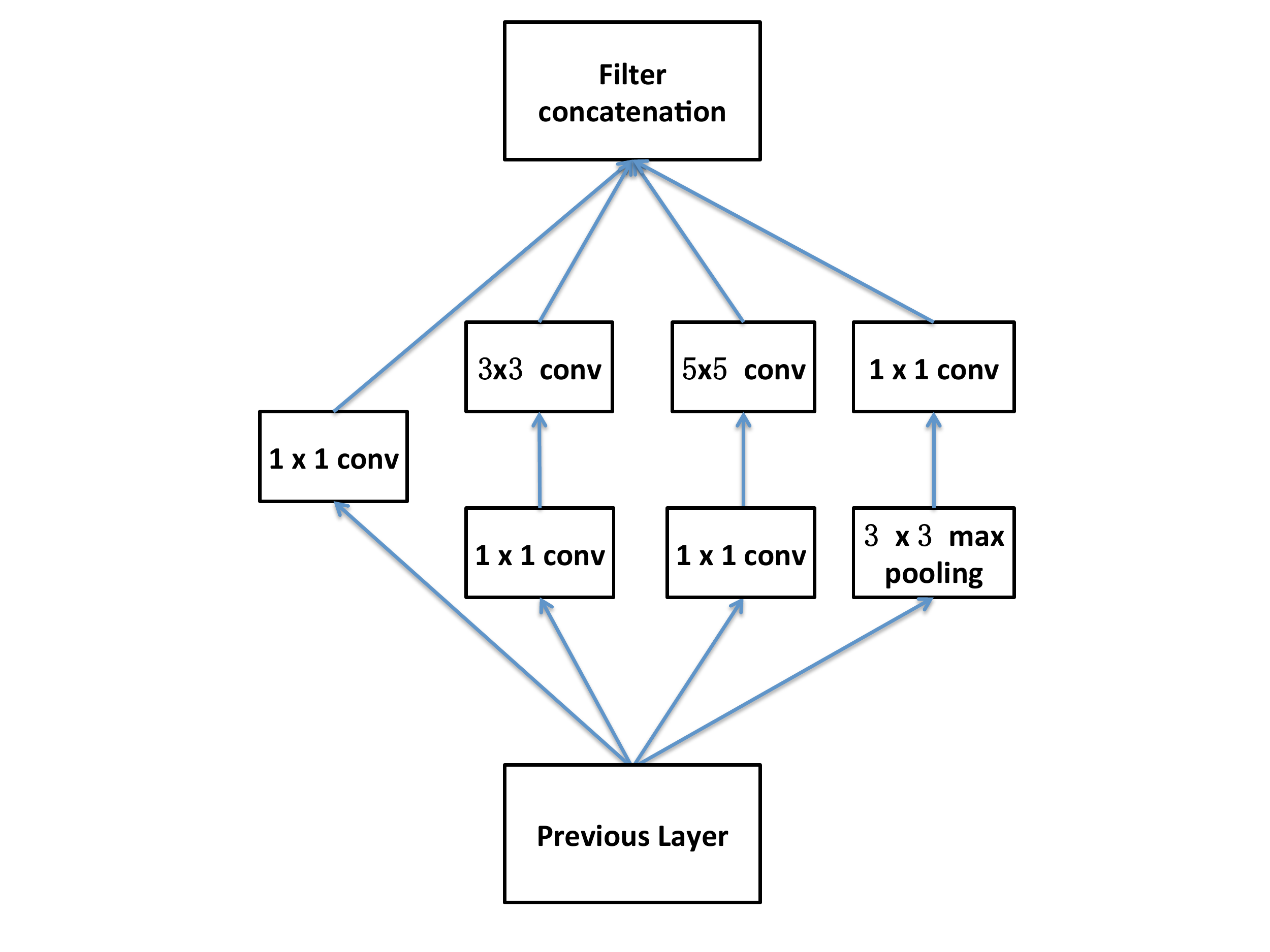}
  \caption{An example of Inception module for GoogLeNet. Notice each convolution are followed by ReLU layer.}\label{fig:inception}
\end{figure}

\subsubsection{VGG Network}

VGG net \citep{simonyan2014very} is a neural network structure  using an architecture with very small ($3 \times  3$) convolution filters, which won the first and the second places in the localization and classification tracks for ImageNet Challenge 2014 respectively. Each block is made by several consecutive $3 \times 3$ convolutions and followed by a max pooling layer. The number of filters for each convolution increases as the network goes deeper. Finally there are three fully connected layers before the softmax transformation.

In this study, we only used VGG net D with 16 layers \citep{simonyan2014very}. We denote it as VGG net for simplicity in the later sections.

\subsubsection{Residual Network}

\begin{figure}[ht]
  \centering
    \includegraphics[width=0.8\textwidth]{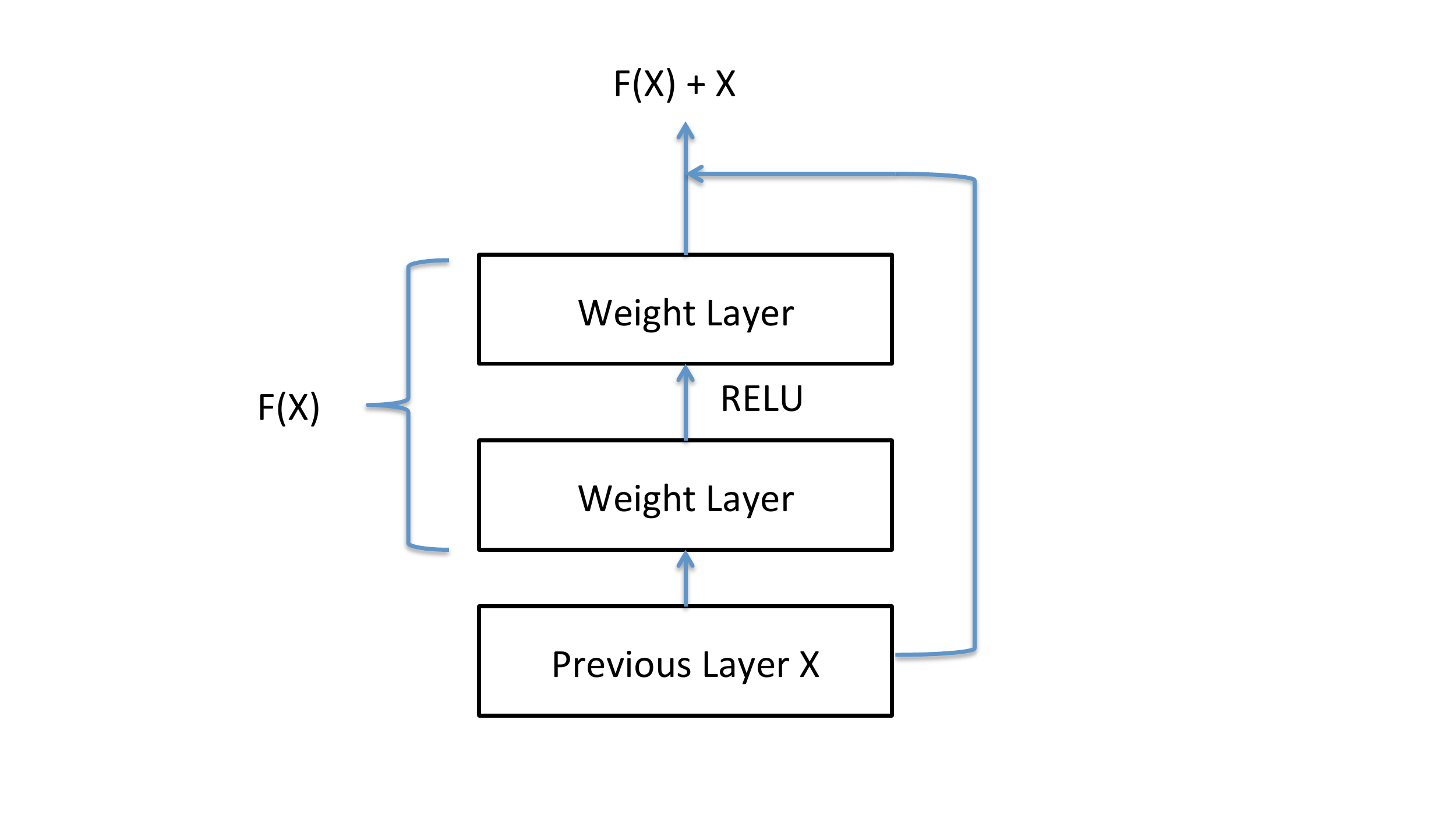}
  \caption{An example of Inception module for GoogLeNet. Notice each convolution are followed by ReLU layer.}\label{fig:res}
\end{figure}

Residual Network \citep{he2015deep} is a network structure that stacked by multiple ``bottleneck'' building blocks. Figure \ref{fig:res} shows an example of so called bottleneck building block, stacked by two regular layer (e.g. convolution layers). In the original study \citep{he2015deep}, each bottleneck building block is made by three convolutional layers, with kernel size 1, 3, and 1. Similar to NIN and GoogLeNet, it uses $1 \times 1$ convolution as dimension reduction to reduce the computation. There is a parameter-free identity shortcut from the starting layer to the final output for each bottleneck block. It solves the degradation problem for deep networks and makes training a very deep neural network possible.

In later sections, we follow the same structure from the original paper for CIFAR-10 data: we use a stack of $6n$ layers with $3 \times 3$ convolutions. The sizes of the feature maps are $\{32, 16, 8\}$ respectively, with $2n$ layers for each feature map size \citep{he2015deep}. There would be $6n+2$ layers including the softmax layer. For example, ResNet with $n=5$ has $32$ layers in total.

\subsection{Training}

For all the models, we split the training data into training (first $4,5000$ images) and validation set (last $5,000$ images). There are 10K testing data.

For the Network-in-Network model, we used Adam with learning rate $0.001$. We followed the original paper \citep{lin2013network}, tuning the learning rate and initialization manually. The training was regularized by $L$-2 penalty with predefined weight $0.001$ and  two dropout layers in the middle of the network, with rate $0.5$.

For VGG net, we slightly modified the training procedure in the original paper \citep{simonyan2014very} for ILSVRC-2013 competitions \citep{zeiler2014visualizing,russakovsky2015imagenet}.  We used SGD with  momentum $0.9$. We started with learning rate $0.01$ and decay divide it by $10$ at every $32k$ iterations.  The training is regularized by $L$-2 penalty with weight $10^{-3}$ and  two dropout layers for the fitst two fully connected layer, with rate $0.5$.

For GoogLeNet, we set base learning rate to be $0.05$, weight decay $10^{-3}$, and momentum $0.9$. We decreased the learning rate by $4\%$ every $8$ epochs. We set the rate to $0.4$ for the dropout layer before the last fully connected layer.

For the Residual Network, we follow the training procedures in the original paper \citep{he2015deep}: we applied SGD with weight decay of $0.0001$ and  momentum of $0.9$. The weight was initialized following the method in \citep{he2015delving}, and we applied batch normalization \citep{ioffe2015batch} without dropout. Learning rate started with $0.1$, and was divided by $10$ at every $32k$ iterations. We trained the model with $200$ epochs.

All the networks were trained with mini-batch size $128$ for 200 epochs.

\subsection{Results}

In this section, we compare the empirical performance for all the ensemble methods we mentioned before, including: Unweighted Averaging (before/after softmax layer), Majority Voting, Bayes Optimal Classifier, Super Learner (with negative log-likelihood loss). We also include discrete SL, with negative log-likelihood loss and 0-1 error loss.. For comparison, we list the base learner which achieved best performance on the \emph{testing set}, as an empirical oracle. 

\subsubsection{Ensemble of Same Network with Different Training Checkpoints}

\begin{table}[ht]
    \caption{Left: Prediction accuracy on the testing set for ResNet 8 trained by 80, 90, 100, 110 epochs. Right:  Prediction Accuracy on the testing set for ResNet 110 trained by 70, 85, 100, 115 epochs.}
  \label{table:diff-epoch-1}
  \parbox{.3\linewidth}{
    \centering
    \begin{tabular}{ |l|r| }
      \hline
      Training Epoch & Prediction Accuracy \\ \hline
      70 & 0.7790 \\\hline
      80 & 0.8245 \\\hline
      90  & 0.8197 \\\hline
      100 & 0.8659 \\\hline
    \end{tabular}
  }
\hspace*{8em}
  \parbox{.3\linewidth}{
    \centering
    \begin{tabular}{ |l|r| }
      \hline
      Training Epoch& Prediction Accuracy \\ \hline
      70 & 0.8896 \\\hline
      85 & 0.8999 \\\hline
      100  & 0.9318 \\\hline
      115 & 0.9354 \\\hline
    \end{tabular}
  }
\end{table}

\begin{table}[ht]
  \centering
   \caption{Prediction accuracy on the testing set for ResNet 8 and 110}
  \label{table:diff-epoch-2}
  \begin{tabular}{ |l|r|r| }
    \hline
    Ensemble               & ResNet 8 & ResNet 110 \\ \hline
    Best Base Learner            &  0.8659  &  0.9354 \\\hline
    SuperLearner           &  $\bold{0.8679}$ & $\bold{0.9358}$ \\\hline
    Discrete SuperLearner (nll)  & 0.8659  & 0.9354 \\\hline
    Discrete SuperLearner (error) & 0.8659  & 0.9354 \\\hline
    Unweighted Average (before softmax)  & 0.8611  & 0.9354 \\\hline
    Unweighted Average  (after softmax)  & 0.8614  & 0.9354 \\\hline
    BOC (before softmax)    &  0.8659  & 0.9318 \\\hline
    BOC (after softmax)     &  0.8659  & 0.9318 \\\hline
    Majority Voting           & 0.8485 & 0.9319 \\\hline
  \end{tabular} 
\end{table}

Table \ref{table:diff-epoch-1} shows the prediction accuracy for the ResNet 8 and 110 after different epochs. As ResNe 8 is much shallower, thus more adaptive during training, we set the smaller interval with epoch 10. Notice there is a great accuracy improvement around epoch 100, due to the learning rate decay.

For ResNet 8, the SL is substantively better than naive averaging and majority voting. Earlier stage learners would have worse performance, which causes the deterioration of the performance for naive averaging. The performance of majority voting is even worse than the best base learner, as the majority of the base learners are under-optimized.

For ResNet 110, the performance for all the meta-learners is similar. One possible explanation is that deeper network is more stable during training.

In this experiment, the weights of BOCs  are dominated by one model, which gives the best performance on the validation set. Thus the BOC is equivalent to the discrete Super Learner with negative likelihood as loss function. In the experiments, BOC performed only as well as the best base learner. In the subsequent experiments, all the BOCs showed the similar dominated weight pattern. Given the practical equivalence with the discrete Super Learner, we don't elaborate further on BOCs, and we will report only the discrete Super Learner's performance.

\subsubsection{Ensemble of Same Network Trained Multiple Times}

Unlike other conventional machine learning algorithms, deep neural networks solve a high-dimensional non-convex optimization problem. Mini-batch stochastic gradient descent with momentum is commonly used for training. Due to non-convexity, networks with same structure but different initialization and training vary a lot. \citep{choromanska2015loss} studied the distribution of loss on the testing set for a certain network structure trained multiple times with SGD. It shows the distribution of loss is more concentrated for deeper neural network. This suggest deep neural networks are less sensitive to randomness in the initialization and training. If so,  ensemble learning would be less helpful for the deeper nets.

To help understand this property, we trained 4 ResNet with 8 layers and 4 ResNet with 110 layers.

\begin{table}[ht]
    \caption{Prediction Accuracy on the testing set for ResNet with 8 and 110 layers}
  \label{table:resnet-models}
  \parbox{.3\linewidth}{
    \centering
    \begin{tabular}{ |l|r| }
      \hline
      Model & Prediction Accuracy \\ \hline
      ResNet 8  0 & 0.8785 \\\hline
      ResNet 8   1 & 0.8819 \\\hline
      ResNet 8   2 & 0.8758 \\\hline
      ResNet 8  3 & 0.8761 \\\hline
    \end{tabular}
  }
\hspace*{8em}
  \parbox{.3\linewidth}{
    \centering
    \begin{tabular}{ |l|r| }
      \hline
      Model & Prediction Accuracy \\ \hline
      ResNet 110   0 & 0.9399 \\ \hline
      ResNet 110   1 & 0.9364 \\\hline
      ResNet 110   2 & 0.9349 \\\hline
      ResNet 110   3 & 0.9395 \\\hline
      
    \end{tabular}
  }
\end{table}

\begin{table}[ht]
  \centering
    \caption{Prediction accuracy on the testing set for ensemble methods. The algorithm candidates are the  ResNets with same structure but trained several times, where the differences come from randomized initialization and SGD. }
  \label{table:same-net}
  \begin{tabular}{ |l|r|r| }
    \hline
  Ensemble                & ResNet 8 &  ResNet 110\\ \hline
    Best Base Learner       & 0.8820  & 0.9399 \\\hline
    SuperLearner            & $\bold{0.9073}$ & $\bold{0.9542}$ \\\hline
    Discrete SuperLearner (nll)  & 0.8820 & 0.9395 \\\hline
    Discrete SuperLearner (error) & 0.8761 & 0.9395 \\\hline
    BOC  (before Sotmax)    & 0.8820 & 0.9395 \\\hline
    BOC  (after Sotmax)     & 0.8820 & 0.9395 \\\hline
    Unweighted Average (before Sotmax)          & 0.9068 & 0.9542 \\\hline
    Unweighted Average (afterbefore Sotmax)     & 0.9068 & 0.9541 \\\hline
    Majority Vote                               & 0.9000 & 0.9510 \\\hline
  \end{tabular}
\end{table}

We trained 4 networks for ResNet 8 and 110 respectively. Table \ref{table:resnet-models} shows the performance of the networks. We further studied the performance of all the meta-learners. Shallow networks enjoyed more improvement ($2.54\%$) compared to deeper networks $1.43\%$ after ensembled by the Super Learner. Due to the similarity of the models, the SL did not show great improvement compared to naive averaging. Similarly, majority voting did not work well, which might also be due to the diversity of the base learners.  The discrete SL with negative log-likelihood loss successfully selected the best single learner in the library, while the  discrete SL with error loss selected a slightly weaker one. This suggests that for finite samples, the Super Learner using the negative log likelihood loss performs better w.r.t. prediction accuracy, than the Super Learner that uses prediction accuracy as criterion.



\subsubsection{Ensemble of Networks with Different Structure}

In this section, we studied  ensemble of networks with different structure. We trained NIN, VGG,and ResNet with 32, 44, 56, 110 layers. Table \ref{table:diff-net} shows the performance of each net on the testing set.

\begin{table}[ht]
  \centering
    \caption{Prediction Accuracy on the testing set for networks with  different structure}
  \label{table:diff-net}
  \begin{tabular}{ |l|r| }
    \hline
    Model     & Prediction Accuracy \\ \hline
    NIN       & 0.8677 \\\hline
    VGG       & 0.8914 \\\hline
    ResNet 32 & 0.9181 \\\hline
    ResNet 44 & 0.9243 \\\hline
    ResNet 56 & 0.9272 \\\hline
    ResNet 110& 0.9399 \\\hline
  \end{tabular}
\end{table}

\subsubsection{Over-confident Model}

\begin{table}[ht]
  \centering
    \caption{Cross-entropy on the testing set for Networks with  different structure}
  \label{table:cross-entropy}
  \begin{tabular}{ |l|r| }
    \hline
    Model     & Cross-entropy \\ \hline
    NIN       & 0.5779 \\ \hline
    VGG       & 1.5649 \\ \hline
    ResNet 32 & 1.5442 \\ \hline
    ResNet 44 & 1.5341 \\ \hline
    ResNet 56 & 1.5327 \\ \hline
    ResNet 110& 1.5242 \\ \hline
  \end{tabular}
\end{table}

As the $0-1$ loss for classification is not differentiable, cross-entropy loss is commonly used as surrogate loss in neural network training. We could see from table \ref{table:cross-entropy} that the cross-entropy is usually negatively correlated with the prediction accuracy. However, we could see that Network-in-Network model has much lower cross-entropy loss compared to all the other models, while it gives worse prediction accuracy. This due to its prediction behavior: we look at the predicted probability of the true labels for the images in the testing set:

\begin{table}[ht]
  \centering
    \caption{Cross-entropy on the testing set for networks with  different structure}
  \label{table:pred-prob}
  \begin{tabular}{ |l|r|r|r|r|r| }
    \hline
    Model     & Image 1 & Image 2  & Image 3  & Image 4  & Image 5\\ \hline
    NIN       & 0.9999 & 0.9999 &  0.09985 &  0.5306 &  1.000  \\\hline
    VGG       & 0.2319  & 0.2319 & 0.2319  & 0.2302 & 0.2314 \\\hline
    ResNet 32 & 0.2319 &  0.2318 & 0.2317 & 0.2316 & 0.2317 \\\hline

  \end{tabular}
\end{table}

It is interesting to observe the high-confidence phenomenon for the Network-in-Network model, where most of the predictions are made with high confidence (predicted probability). Such high-confident networks usually achieve much smaller surrogate loss (negative log-likelihood loss in our example) on the testing set, but not necessary smaller 0-1 error loss. Though all the networks suffered from over-fitting, only the NIN net showed the over-confidence. In addition,  NIN has higher training cross-entropy loss ($0.13104$) compared to VGG ($0.02233$). Thus it is not reasonable to blindly attribute the over-confidence to the over-fitting.

When several base learners suffer from the over-confidence issue, the performance of model averaging would be seriously deteriorated: the unweighted average score/probability would be dominated by the over-confident models. When all the models are over-confident, the unweighted average is identical to the majority vote.

In addition, the VGG net and the ResNet with 32 layers had very similar predicted probability, even though their structure is totally different (agree on first 3 digits on most observations). However, this special pattern is beyond the scope of this study.

We empirically study the impact of over-confident network candidates for  ensemble methods:
we have five candidates in the ensemble library: NIN, VGG, ResNet 32, ResNet 44, and ResNet 56. We compare the performance with/without adding NIN, which is the only over-confident net.

\begin{table}[ht]
  \centering
    \caption{Prediction accuracy on the testing set for ensemble methods. The algorithm candidates include  NIN, VGG, ResNet 32, ResNet 44, and ResNet 56. We compare the performance with/without the over-confident NIN network.}
  \label{table:over-confident}
  \begin{tabular}{ |l|r|r| }
    \hline
    Ensemble                & Without NIN &  With NIN\\ \hline
    Best Base Learner   & 0.9399 & 0.9399 \\ \hline
    SuperLearner            & $\bold{0.9469}$ & $\bold{0.9475}$\\\hline
    Discrete SuperLearner (nll)  & 0.9399 & 0.8677\\\hline
    Discrete SuperLearner (error)  & 0.9399 & 0.9399\\\hline
    BOC (before softmax)      & 0.9399 & 0.8677\\\hline
    BOC (after  softmax)      & 0.9399 & 0.8677\\\hline
    Unweighted Average (before softmax)      & 0.9456 & 0.9223\\\hline
    Unweighted Average (after softmax)       & 0.9455 & 0.8974\\\hline
    Majority Vote           & 0.9433 & 0.9413\\\hline
  \end{tabular}
\end{table}

Table \ref{table:over-confident} shows the performance of the ensemble algorithms on the testing set. The unweighted average model was weakened by the NIN net: over-confidence made NIN dominate the others, and led to $0.23\%$ (before softmax) and $5\%$ (after softmax) decrease in the prediction accuracy. The naive average before softmax was less influenced as the scale of networks are different. The majority vote algorithm was not influenced too much by the extra candidate, which is not surprising. The over-confident network only weakened discrete SL with negative log-likelihood loss, while did not influence the discrete SL with error loss. The Super Learner successfully harnessed the over-confident model: adding NIN helped increase the prediction accuracy from $0.9405$ to $0.9414$.

\subsubsection{Learning from Weak Learner}

We hope our ensemble method could learn from all the models, even though there might be base learners with weaker overall performance compared to the other learners in the library. In this experiment, we used under-trained GoogLeNets \citep{szegedy2015going} as the weak candidates. The original paper \citep{szegedy2015going} did not describe explicitly how to automatically train/tune the network in CIFAR 10 data set. We  set the initial learning rate to be $0.05$, with momentum $0.96$, and decreased the learning rate by $4\%$ every 8 epochs. This did not give satisfactory performance: the prediction accuracy on the testing set is around $0.83$. To avoid the impact of over-confidence, we removed the NIN net. Thus the weakest base learner in the library is the VGG net, which achieved $0.8914$ accuracy on the testing set. We observe that the difference in prediction accuracy for the VGG net and the GoogLeNet is around $6\%$, which means our GoogLeNet model is substantially weaker than other candidates.

We trained the GoogLeNet 5 times and then compare the performance of different ensemble methods with/without such 5 googLeNets in the library. 

\begin{table}[ht]
  \centering
    \caption{Prediction accuracy on the testing set for ensemble methods. The algorithm candidates include  VGG, ResNet 32, ResNet 44, and ResNet 56. We compared the performance with/without five under-optimized GoogLeNets.}
  \label{table:weak}
  \scalebox{0.75}{
  \begin{tabular}{ |l|r|r|r| }
    \hline
   Ensemble                 & Without GoogLeNet &  With 3 GoogLeNets & With 5 GoogLeNets\\ \hline
      Best Base  Learner  & 0.9399 & 0.9399 & 0.9399\\ \hline
    SuperLearner             & $\bold{0.9475}$ & $\bold{0.9477}$ & $\bold{0.9477}$ \\\hline
    Discrete SuperLearner (nll)    & 0.9399 & 0.9399 & 0.9399 \\\hline
    Discrete SuperLearner  (error)  & 0.9399 & 0.9399 & 0.9399 \\\hline
    BOC   (before softmax)   & 0.9399 & 0.9399 & 0.9399 \\\hline
    BOC   (after softmax)    & 0.9399 &  0.9399 & 0.9399 \\\hline
    Unweighted Average    (before softmax)   & 0.9456 & 0.9326 & 0.9001 \\\hline
    Unweighted Average   (after softmax)     & 0.9455 & 0.9329 & 0.9007 \\\hline
    Majority Vote                            & 0.9433 & 0.9263 & 0.8720 \\\hline
  \end{tabular}
  }
\end{table}

In the experiment, adding many weaker candidates  deteriorated the performance of the unweighted average. The majority voting was slightly influenced when there were  only few weak learners, while would be dominated if the number of the weak learner was large. Unweighted averaging also failed in this case. BOCs remained unchanged as  the  likelihood on the validation set is still dominated by the same base learner. Super Learner shows exciting success in this setting: the prediction accuracy remained stable with the extra weak learning.

\subsubsection{Prediction with All Candidates}

 As the number of base learners is usually much smaller than the sample size and there is usually no apriori which learner would achieve best performance, it is encouraged to apply as rich library as possible to improve the performance of Super Learner. In this experiment, we simply put all the networks mentioned before into the library of all the ensemble methods.

\begin{table}[ht]
  \centering
    \caption{Prediction accuracy on the testing set for all the ensemble methods using all the networks mentioned in this study as base learners.}
  \label{table:all}
  \begin{tabular}{ |l|r| }
    \hline
    Ensemble                & Accuracy        \\ \hline
    Best base learner       & 0.9399          \\ \hline
    SuperLearner            & $\bold{0.9502}$ \\\hline
    Discrete SuperLearner (nll)   & 0.9395         \\\hline
    Discrete SuperLearner (error) & 0.9395         \\\hline
    BOC (before softmax)      & 0.9395        \\\hline
    BOC (after  softmax)      & 0.9395        \\\hline
    Unweighted Average (before softmax)  & 0.9444 \\\hline
    Unweighted Average (after softmax)   & 0.9448 \\\hline
    Majority Vote             & 0.9410        \\\hline
  \end{tabular}
\end{table}

Table \ref{table:all} shows the performance of all the ensemble methods as well as the base learner with the best performance. Due to the large proportion of weak learners (e.g. under-fitted GoogLeNet, and the networks trained with less iterations in the first experiment) and  the over-confident learners (NIN), all the other ensemble methods have much worse performance compared to Super Learner. This is another strength of the Super Learner: by simply putting all the potential base learners  into the library, the Super Learner  computes the weights data-adaptively, which does not require any tedious pre-selecting procedure based on human experience.

\subsection{Discussion}

We studied the relative performance for several widely used ensemble methods with deep convolutional neural networks as base learners on the CIFAR 10 data set, which is a commonly used benchmark for image classification. The unweighted averaging proved surprisingly successful when the performance of the base learners are comparable. It outperformed majority voting in almost all the experiments. However, the unweighted averaging is proved to be sensitive to over-confident candidates. The Super Leaner addressed this issue by simply optimizing a weight on the validation set in a data-adaptive manner. This ensemble structure could be considered as a $1 \times 1$ convolution layer stacked on the output of the base learners. It could adaptively assign weight on base learners, which enables weak learner to improve the prediction.

Super Learner is proposed as a cross-validation based ensemble method. However, since CNN are computationally intensive and that validation sets are typically large in image recognition tasks, we used the validation set of the neural networks for computing the weights of Super Learner(single-split cross-validation), instead of using  conventional cross validation (multiple-fold cross-validation). The structure is simple and could be easily extended. One potential extension of the linear-weighted Super Learner would be stacking several $1 \times 1$ convolutions with non-linear activation layers in between. This structure could mimic the cascading/hierarchical ensemble \citep{wang2014cascaded,su2009hierarchical}. Due to the small number of parameters, we hope this meta-learner would not overfit the validation set and thus would help improve the prediction. However this involves non-convex optimization and the results might not be stable. We leave this as future work.


\bibliographystyle{abbrvnat} 
\bibliography{CNN-SL}

\end{document}